\pdfoutput=1

\documentclass[11pt]{article}

\usepackage[final]{coling}

\usepackage{times}
\usepackage{latexsym}

\usepackage[T1]{fontenc}

\usepackage[utf8]{inputenc}

\usepackage{microtype}

\usepackage{inconsolata}

\usepackage{graphicx}

\usepackage{multirow}
\usepackage{graphicx}
\usepackage{booktabs}
\usepackage{xcolor}
\usepackage{colortbl}
\usepackage{array}
\usepackage{amsmath}
\usepackage{amssymb}
\usepackage{xspace} 
\usepackage{enumitem}

\usepackage{algorithm}
\usepackage{algorithmic}

\newcommand*{\modelname}{\text{
MKP-QA
}\@\xspace}

\title{
Federated Retrieval Augmented Generation for Multi-Product Question Answering
}

\author{
    Parshin Shojaee\textsuperscript{\rm 1}\thanks{Work done while interning at Adobe}, \
    Sai Sree Harsha\textsuperscript{\rm 2}, \ 
    Dan Luo\textsuperscript{\rm 2}, \
    Akash Maharaj\textsuperscript{\rm 2}, \
    Tong Yu\textsuperscript{\rm 2}, \
    Yunyao Li\textsuperscript{\rm 2} \\
    \textsuperscript{\rm 1}Virginia Tech \hspace{5mm}
    \textsuperscript{\rm 2}Adobe \\
    parshinshojaee@vt.edu, 
    \{ssree, dluo, maharaj, tyu, yunyaol\}@adobe.com
  }

\begin{document}
\maketitle
\begin{abstract}
Recent advancements in Large Language Models and Retrieval-Augmented Generation have boosted interest in domain-specific question-answering for enterprise products. 
However, AI Assistants often face challenges in multi-product QA settings, requiring accurate responses across diverse domains.
Existing multi-domain RAG-QA approaches either query all domains indiscriminately, increasing computational costs and LLM hallucinations, or rely on rigid resource selection, which can limit search results.
We introduce \modelname, a novel multi-product knowledge-augmented QA framework with probabilistic federated search 
across domains and relevant knowledge.
This method enhances multi-domain search quality 
by aggregating query-domain and query-passage probabilistic relevance. 
To address the lack of suitable benchmarks for multi-product QAs, we also present new datasets focused on three Adobe products: Adobe Experience Platform, Target, and Customer Journey Analytics.
Our experiments show that 
\modelname
significantly
boosts multi-product RAG-QA performance in terms of both retrieval accuracy and response quality.
\end{abstract}

\section{Introduction}
\vspace{-0.2em}
The rapid advancement of Large Language Models (LLMs) and Retrieval-Augmented Generation (RAG) has sparked significant interest in question-answering (QA) systems for domain-specific applications and products. 
This technology has significantly enhanced enterprise product support \cite{sharma2024retrieval}, offering users more efficient and accurate ways to access information about complex product ecosystems with specific details, terminologies, usage procedures as well as related use cases. 
However, as the complexity of enterprise software suites grows, so does the challenge of providing accurate and comprehensive answers to user queries that may span multiple products or require cross-product knowledge.

In the context of enterprise product-related QA tasks, users often need to navigate multiple products and understand how they can be integrated to address specific use cases. This multi-product and cross-product nature of queries presents unique challenges for traditional RAG-QA approaches, particularly in context augmentation from diverse knowledge resources. These challenges are especially pronounced in industrial settings, where the accuracy of information retrieval and response generation directly impact customer satisfaction.


Current approaches to multi-domain search in RAG-QA systems typically fall into two main categories: (1)~querying all product domains indiscriminately \citep{wu2024multi}, or (2)~employing resource selection techniques \citep{wang2024feb4rag,wang2024resllm}. Both methods have significant drawbacks. The first approach, while comprehensive, can lead to increased computational costs and even potentially compromise answer quality due to the higher likelihood of LLM hallucination or inaccurate responses when presented with diverse and irrelevant concepts from different domains. 
The second approach, which attempts to narrow the search to specific domains, risks propagating selection errors that can limit the scope of the search and potentially miss crucial cross-product information, leading to incomplete or misleading answers in complex enterprise scenarios.

To address these challenges, we propose \textbf{\modelname}, a novel \textbf{M}ulti-domain \textbf{K}nowledge-augmented \textbf{P}roduct RAG-\textbf{QA} framework that optimizes multi-domain question answering. \modelname is designed to meet the specific needs of enterprise software ecosystems, where accurate cross-product information retrieval is essential. The core of \modelname employs a federated search mechanism \citep{fedsearch} that intelligently navigates across multiple product domains and their associated relevant corpus. This approach allows \modelname to search across a diverse range of enterprise products and their associated documentation without the need to centralize all information into a single, monolithic database – a significant advantage in large-scale enterprise deployments.


The complex nature of multi-product QA in enterprise settings necessitates a more nuanced approach than simple federated search. Cross-product queries often require information from multiple domains with overlapping terminologies, including less obvious ones.
To address these practical needs, we enhance \modelname's federated search with a probabilistic gating mechanism, serving three crucial functions:
($i$)~\textit{Exploration-Exploitation Balance}, enabling both exploitation of known relevant domains and exploration of less obvious ones, crucial for cross-product queries; ($ii$)~\textit{Error Mitigation}, using likelihoods in domain selection to safeguard against misclassification and missed information in domain router; and ($iii$)~\textit{Adaptive Query Processing}, allowing flexible and context-aware searching.
By aggregating query-domain and query-document relevance scores through this mechanism, \modelname enhances multi-domain document retrieval for RAG-QA systems. This adaptation of federated learning techniques \citep{ashman2022partitioned,huang2022stochastic} to our specific challenges enables more accurate, cross-domain product knowledge integration, particularly valuable in complex enterprise software ecosystems.
To address the lack of suitable multi-product QA benchmarks, we also introduce new benchmark datasets focused on three Adobe products: Adobe Experience Platform (AEP), Adobe Target, and Adobe Customer Journey Analytics (CJA).
These datasets, which we intend to release publicly, consist of user queries and corresponding documents from Adobe product documentation. They serve as valuable resources for evaluating domain-specific and cross-domain RAG-QA systems across product domains. The datasets will be made available pending Adobe's approval.
 

 


Our experimental findings demonstrate significant improvements in the accuracy of multi-product question answering. 
By introducing new benchmark datasets and proposing an innovative framework, we seek to push the boundaries of AI Assistants in product-related QA. 
Importantly, \modelname achieves this without requiring separate domain-specific LLM fine-tuning or the training of adaptive modules across various product domains.




\section{Related Work}

\vspace{-0.2em}
\subsection{Domain-specific Question-Answering}
\vspace{-0.2em}
Domain-specific QA has seen significant advancements across various fields, addressing the unique challenges posed by specialized knowledge and terminology. Research efforts have focused on developing tailored methods and datasets for domains such as biomedical \cite{biomedicalQA1}, physics \cite{physicsQA1}, finance \cite{financeQA1}, and legal \cite{legalQA1}. These works have contributed to improving QA accuracy and relevance within their respective fields.
In the context of product-related QA, which is most relevant to our work, efforts have been more limited. 
Notable among these is the dataset in \cite{geeval} which focuses on Microsoft product queries. However, this dataset primarily consists of yes/no questions, with only a small portion requiring more complex generative text answers.
Our work extends this line of research by addressing multi-product QA in enterprise software ecosystems. We focus on more complex, cross-domain queries that often require integrating knowledge from multiple products - a scenario common in enterprise settings but underexplored in current literature.


\vspace{-0.2em}
\subsection{Retrieval Augmented Generation}
Retrieval augmented generation (RAG) has recently emerged as a powerful approach for enhancing the performance of LLMs in knowledge-base QA tasks. RAG combines the strengths of retrieval-based and generation-based methods to produce more accurate and faithful responses.
\cite{lewis2020retrieval} introduced the foundational RAG model, which retrieves relevant documents and conditions its output on both the retrieved information and the input query. Subsequent works have further improved it with \cite{guu2020retrieval} developing REALM for joint training of retriever and generator, and \cite{karpukhin-etal-2020-dense} introducing dense passage retrieval for improved efficiency.
Recent research has explored RAG in domain-specific contexts. \cite{head2021augmenting} adapted RAG for scientific literature, while \cite{khattab2022demonstrate} investigated its application in customer support settings.
Our work extends this line of research
by introducing a novel multi-domain RAG framework that addresses the specific challenges of enterprise systems, where queries often span multiple products and require integration of diverse knowledge.


\subsection{Multi-domain Document Retrieval}
Multi-domain document retrieval presents unique challenges, particularly in enterprise product settings where information is often distributed across diverse and overlapping knowledge sources. Research in this area has focused on developing methods to accurately retrieve relevant information from multiple sources.
Federated search approaches \cite{fedsearch} enable querying multiple distributed indexes simultaneously, while domain adaptation techniques \cite{neural-domainadap} handle transfer across diverse search domains. 
Recent work leveraging LLMs for retrieval resource selection \cite{wang2024resllm} has also shown strong zero-shot performance. 
Our work extends these efforts by introducing a stochastic gating mechanism combined with federated search,
tailored for RAG-QA pipelines in complex environments with cross-product queries.

\begin{figure*}[t]
\centering
\includegraphics[width=0.8\textwidth]{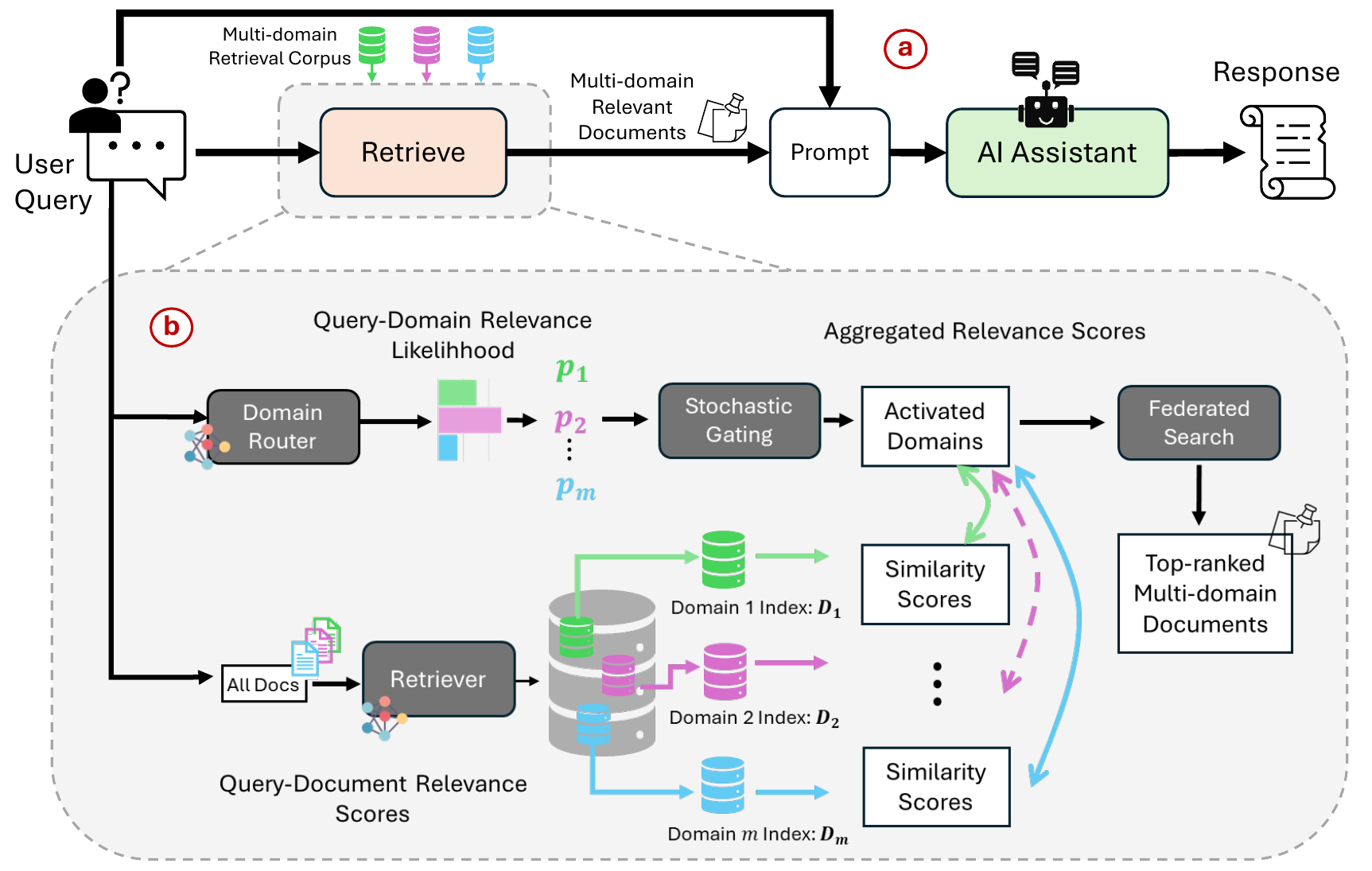}
\caption{
\textbf{Overview of the \modelname framework.} 
\textbf{\textcircled{a}} The main RAG-QA pipeline: retrieval of multi-domain documents, prompt augmentation, and response generation. 
\textbf{\textcircled{b}} Detailed view of the multi-domain knowledge augmentation: 
A domain router estimates query-domain relevance while a retriever finds relevant documents across product domains. A stochastic gating mechanism determines active domains, for which query-domain and query-document relevance scores are aggregated into a unified ranking of multi-domain documents. These top-ranked documents then augment the prompt,
enabling effective cross-domain product QA.
}
\label{fig:mkpqa-main}
\end{figure*}

\vspace{-0.2em}
\section{Methodology}
Our \modelname framework, shown in Fig.~\ref{fig:mkpqa-main}, integrates components for domain relevance, exploration-exploitation, retrieval, and multi-domain aggregation, detailed below.

\vspace{-0.2em}
\subsection{Query-Domain Router}
To effectively estimate the query-domain relevance scores.
we leverage a query-domain router $\mathcal{F}:Q\rightarrow \left[0,1\right]^m$, mapping from 
the space of queries $Q$ to the $m$ product domains as a multi-label classification task. We use a Transformer model \cite{vaswani2017attention}, specifically a variant of BERT, fine-tuned for our multi-domain classification task: 
$\mathcal{F}(q) = \sigma(W+\text{BERT}(q) + b)$,
where $\text{BERT}(q)$ is the contextualized representations of query $q$ at \texttt{[cls]} token;
$W$ and $b$ are learnable 
parameters;
and $\sigma$ is the sigmoid activation function. 
As this is a multi-label classification task, we employ a binary cross-entropy loss for each domain, summed over all domains.
At inference, we estimate the query-domain relevance likelihood with the trained domain router: $\mathcal{F}(q)=\left[p_1, p_2, \ldots, p_m\right]$.

\vspace{-0.2em}
\subsection{Stochastic Gating}
To address the challenge of balancing exploitation of high-confidence domains with exploration of potentially relevant but less certain domains, we introduce a stochastic gating mechanism with adaptive threshold control. This approach allows for dynamic adjustment of the search space based on the Query-Domain Router's confidence and the inherent uncertainty in multi-domain RAG-QA.
We define an adaptive threshold $\tau(q)$ for query $q$ based on the entropy of the domain probability distribution $p$: 
$\tau(q)=\tau_0 ( 1-\frac{-\sum^m_j{p_j \log{(p_j)}}}{\log{(m)}} )$,
where $\tau_0$ is the base threshold hyperparameter; 
and $[p_1,\ldots,p_m]$ is the vector of domain probabilities output by the 
router. 
We utilizie stochastic gating function $\mathcal{G}:M\times Q \rightarrow \{0,1\}$, with $M$
as the domain space and $Q$ as the query space, to facilitate domain selection and introduce exploration. This function is defined as
$\mathcal{G}(q,j)=\text{Bernouli}\left(\min{\left(1,p_j/\tau(q)\right)}\right)$,
where $\mathcal{G}(q,j)$ is the domain $j$-th selection for query $q$ based on the Bernouli sampling.


\subsection{Query-Document Retriever}
To facilitate efficient and effective retrieval of relevant documents across multiple product domains, we employ a bi-encoder architecture for our Query-Document Retriever. This model generates dense vector representations for both queries and documents, enabling rapid similarity computations in the embedding space.
Our retriever embedding model $E$ is based on the Sentence-BERT \cite{reimers2019sentence}
with shared weights for query and document encodings, generating dense vector representations $E_{\theta}(q)$ and $E_{\theta}(d)$ for query $q$ and document $d$.

We fine-tune the retriever model on our multi-domain dataset using a contrastive learning approach with a symmetric supervised variant of the InfoNCE loss \cite{InfoNCE-2018}, 
incorporating supervised relevance labels and symmetry, which we find particularly effective for query-document retrieval tasks in multi-domain settings. The retriever loss function is
$\mathcal{L}_{r}=-(\mathcal{L}_{q2d}+\mathcal{L}_{d2q})/2$,
where $\mathcal{L}_{q2d}$ and $\mathcal{L}_{d2q}$ represent the query-to-document and document-to-query directional losses.
The $\mathcal{L}_{q2d}$ is computed as follows and the $\mathcal{L}_{d2q}$ can be obtained similarly. 

\vspace{-1.0em}
{\scriptsize
$$
\mathcal{L}_{q2d} = \sum_{q}{\sum_{d_+\in\mathcal{D}_+}}{  
  \frac{   {\exp{\left(s(q,d_{+})/\tau\right)}}  }{ {\exp{\left(s(q,d_{+})/\tau\right)}} + \sum_{d\in\mathcal{D}_{-}}{\exp{\left(s(q,d)/\tau\right)}}}     }$$
}

\vspace{-0.2em}
\noindent where $\mathcal{D}_+$ and $\mathcal{D}_-$ are the set of annotated positive and negative document pairs for the given query $q$ within batch; $s(q,d)$ is the dot-product similarity score between query $q$ and document $d$ embedding: $s(q,d)=E(q) \cdot E^T(d)$; and $\tau$ is a temperature hyperparameter.
At inference, we compute the embeddings of all documents in the corpus offline and save in a vector database. For a given query, we compute its embedding and retrieve the top-k documents using similarity score search. 


\vspace{-0.2em}
\subsection{Federated Search}
For a given query $q$, we define the set of active domains $\mathcal{A}(q)=\{j\in M | \ \mathcal{G}(j,q)=1\}$. For each active domain $j\in \mathcal{A}(q)$, we retrieve the top-k documents $D_j=\{d^1_j, \ldots, d^k_j\}$ and their respective query-document relevance scores $S_j = \{s^1_j,\ldots,s^k_j\}$ obtained from the bi-encoder retriever detailed above: $s^i_j = s(q,d^i_j)=E(q)\cdot E^T(d^i_j)$.
Next, we aggregate these to a unified domain-aware retrieval scoring 
$U(j,q,d^i_j)=\mathcal{F}(q)[j] \cdot s(q,d^i_j) = p_j \cdot s^i_j$,
where $U(\cdot)$ is the unified multi-domain ranking score function for query $q$, domain $j$ and retrieved document $d^i_j$ at position $i$ in this domain. 
The final multi-domain ranked set of documents with federated search $D^*$ is obtained by selecting the top-k documents across all the active domains: 
$D^* = \arg\max_k{ \{U(j,q,d^i_j) | j \in \mathcal{A}(q), d^i_j \in D_j\} }$.
These top-ranked documents from multiple domains are then augmented to the prompt and fed to LLM for the product QA.

\section{Dataset Creation and Statistics}

\subsection{Data Sources}
The corpus is derived from the publicly available Adobe Experience League (ExL) documentation\footnote{\url{https://experienceleague.adobe.com/en/docs}}, focusing on three key products: Experience Platform (AEP), Target, and Customer Journey Analytics (CJA). These web-pages provide comprehensive information on product concepts, capabilities, troubleshooting guides, and usage instructions.

\subsection{Data Pre-processing}
The data preparation process involves several steps: \textcircled{1}~\textit{Web Crawling:}~We employ a custom crawling script to extract content from the ExL web-pages. This script navigates through the documentation, capturing textual information while omitting images and converting clickable and in-section links to plain text for consistency.
\textcircled{2}~\textit{Initial Segmentation:}~The extracted content is initially segmented based on HTML header tags. This approach creates distinct sections that typically correspond to specific topics or tasks within each documentation. \textcircled{3}~\textit{Document Chunking:}~To optimize the corpus for efficient retrieval and context preservation, we implement the following chunking strategy. Each web-page is divided at every header level, creating initial chunks that align with the document's logical structure. If any section exceeds a pre-defined token limit (512 tokens), we utilize LangChain's 
hierarchical splitting approach based on a specified character list. This method prioritizes maintaining the integrity of paragraphs, sentences, and words, ensuring that semantically related content remains together as much as possible. 

\subsection{Data Creation}
Our dataset comprises 
query-document pairs from three Adobe product domains (AEP, Target, and CJA). We employed two complementary approaches for data creation: ($i$)~\textit{Subject Matter Expert (SME) Dataset:} Product experts manually created query-document pairs based on respective ExL web-pages for each domain. They wrote queries for documents extracted from web-pages and annotated the relevance of each pair based on their product expertise. ($ii$)~\textit{Synthetic LLM-Assisted (SLA) Dataset:} To ensure comprehensive coverage, we leveraged GPT-4 to generate queries for chunked documents extracted from ExL web-pages. Product experts from each domain subsequently reviewed these query-document pairs to guarantee accuracy and relevance.
For cross-domain data creation, we followed a similar process where product experts first identified ExL web-pages with overlapping documentation across products; GPT-4 then generated queries for these cross-domain web-pages; and product experts reviewed the queries for relevance to the cross-domain documentation content. This approach ensured that positive document pairs per query were designed to span different domains, enhancing the dataset's utility for multi-domain product RAG-QA research.

To enhance the dataset with both positive and challenging negative examples, we employed a systematic approach for document pairing. For each question in the dataset, we utilized two strategies: (1)~pairing the question with other document chunks from the same web-page as the golden document, and (2)~when insufficient negative pairs were available from the original page, sampling document chunks from URLs closely related to the web-page containing the golden document. This method ensures a diverse and representative set of negative examples. Finally, we leveraged GPT-4 to annotate the relevance of each query-document pair. Using the prompt detailed in Appendix (Figure \ref{fig:prompt-annot}), GPT-4 assigns binary labels (Yes/No) to the relevance of all pairs.

\begin{table}[t]
\centering
\renewcommand{\arraystretch}{1.0}
\fontsize{7pt}{9pt}\selectfont
\begin{tabular}{llccc}
\toprule
\multirow{2}{*}{Data Type} & \multirow{2}{*}{Metric} & \multicolumn{3}{c}{Uni-Domain} \\ 
\cmidrule(lr){3-5} 
 &  & AEP & CJA & Target \\ 
\midrule
\multirow{4}{*}{ SME} 
& \# of query-doc pairs & 2,970& 1,035& 521 \\
 & Avg. length of queries & 9.31& 9.75& 9.12 \\
 & Avg. length of docs & 87.59& 207.43& 101.15 \\
 & \% of positive pairs  & 8.95\%& 11.27\%& 10.23\% \\
\midrule
\multirow{4}{*}{ SLA} 
& \# of query-doc pairs & 28,860 & 27,820& 29,610\\
 & Avg. length of queries & 10.75& 11.80& 11.69 \\
 & Avg. length of docs & 143.78& 146.89& 107.15 \\
 & \% of positive pairs  & 17.53\%& 18.28\%& 20.26\% \\
\bottomrule
\end{tabular}
\label{tab:datastat-uni}
\end{table}

\begin{table}[t]
\centering
\renewcommand{\arraystretch}{1.0}
\resizebox{\columnwidth}{!}{%
\fontsize{10pt}{12pt}\selectfont
\begin{tabular}{llccc}
\toprule
\multirow{2}{*}{Data Type} & \multirow{2}{*}{Metric} & \multicolumn{3}{c}{Cross-Domain} \\ 
\cmidrule(lr){3-5} 
 &  & AEP + CJA & AEP + Target & CJA + Target \\ 
\midrule
\multirow{4}{*}{ SLA} 
& \# of query-doc pairs & 880& 1,370& 480 \\
 & Avg. length of queries & 14.70& 14.92& 13.68 \\
 & Avg. length of docs & 141.15& 97.72& 
95.49 \\
 & \% of positive pairs  & 19.21\% & 19.56\%& 18.37\% \\
\bottomrule
\end{tabular}
}
\caption{
Statistics for the Adobe multi-product uni-domain (\textbf{top}), and cross-domain (\textbf{bottom}) RAG datasets
}
\label{tab:datastat-cross}
\end{table}

\subsection{Data Analysis and Statistics}
Our dataset encompasses questions, documents, corresponding source web-page URLs and titles, and annotations across the Adobe AEP, Target, and CJA domains. The questions fall into two main categories: (1)~\textit{"What-is"} or \textit{"Where-is"} questions about product concepts (e.g., "What is a union schema?", "What is an audience?"); and \textit{"How-to"} questions about usage instructions (e.g., steps for "adding services to a datastream" or "looking up a sandbox"). Table~\ref{tab:datastat-cross} provides key data statistics for uni-domain and cross-domain datasets, including the count of question-document pairs, average lengths of questions and documents, and the ratio of positive pairs in the datasets.

\vspace{-0.3em}
\section{Experiments}
\label{sec:exp}
Our experimental study aims to evaluate the effectiveness of \modelname in comparison with various baselines 
for multi-domain RAG-QA
on Adobe datasets. 
We conducted a series of experiments on both uni-domain and cross-domain datasets.
Throughout our experiments, we utilized \texttt{GPT-3.5-turbo-1106} and \texttt{GPT-4-0314} models from Azure OpenAI.


\subsection{Baselines}
\paragraph{Unified Index and Search (UIS):} This baseline uses a single multi-domain index with a retriever fine-tuned on all three product domains. Search is performed across the entire index without considering domain relevance to the query.

\paragraph{Router Filter and Search (RFS):} A domain router selects the most likely domain for each query, limiting the search to documents tagged for that domain within the unified index.

\paragraph{LLM Filter and Search (LFS):} Using the ReSLLM method \cite{wang2024resllm}, this baseline leverages GPT-4 in a zero-shot manner for domain selection, then searches within that domain's subset of the unified index (see Appendix Figure~\ref{fig:prompt-resllm} for prompt details).

In all baselines, vector similarity is used to retrieve the top-$5$ most relevant documents by comparing the query's vector to document vectors. These documents are then augmented into the assistant's prompt for response generation.
\subsection{Evaluation Methods}
To comprehensively assess the performance of our multi-domain RAG-QA pipeline, we employ a variety of evaluation metrics targeting both retrieval accuracy and response quality:
($i$)~\textit{Retrieval Accuracy:} 
For evaluating retrieval performance across multiple domains, we use the Acc@Top1 metric. This metric represents the percentage of queries for which the golden (most relevant) multi-domain documents are correctly retrieved as the top-ranked candidate. Our focus on the first document is motivated by recent RAG studies \cite{liu2024lost,xu2024context}
showing that the top-ranked document, when added to the prompt, most significantly influences the LLM's response.
($ii$)~\textit{Response Quality:} 
To assess the quality of generated answers for product QA, we employ Relevancy and Faithfulness analysis.
In the former, 
we incorporate GPT-4, following the prompting strategy in \cite{zheng2024judging},
to evaluate the relevancy and helpfulness of the 
generated responses to queries. 
Given the domain-specific nature of product QA, we also utilize the RAGAS\footnote{\url{https://docs.ragas.io}} framework \cite{es2023ragas} to assess the faithfulness of generated responses. In this metric, we decompose each response into individual statements and cross-check them against the ground truth documentation for each query with the help of GPT-4. The faithfulness score is computed as the percentage of statements that GPT-4 recognizes can be directly inferred from the provided context. Detailed prompts for these metrics are available in the Appendix.


\begin{figure}[t]
\centering
\includegraphics[width=\columnwidth]{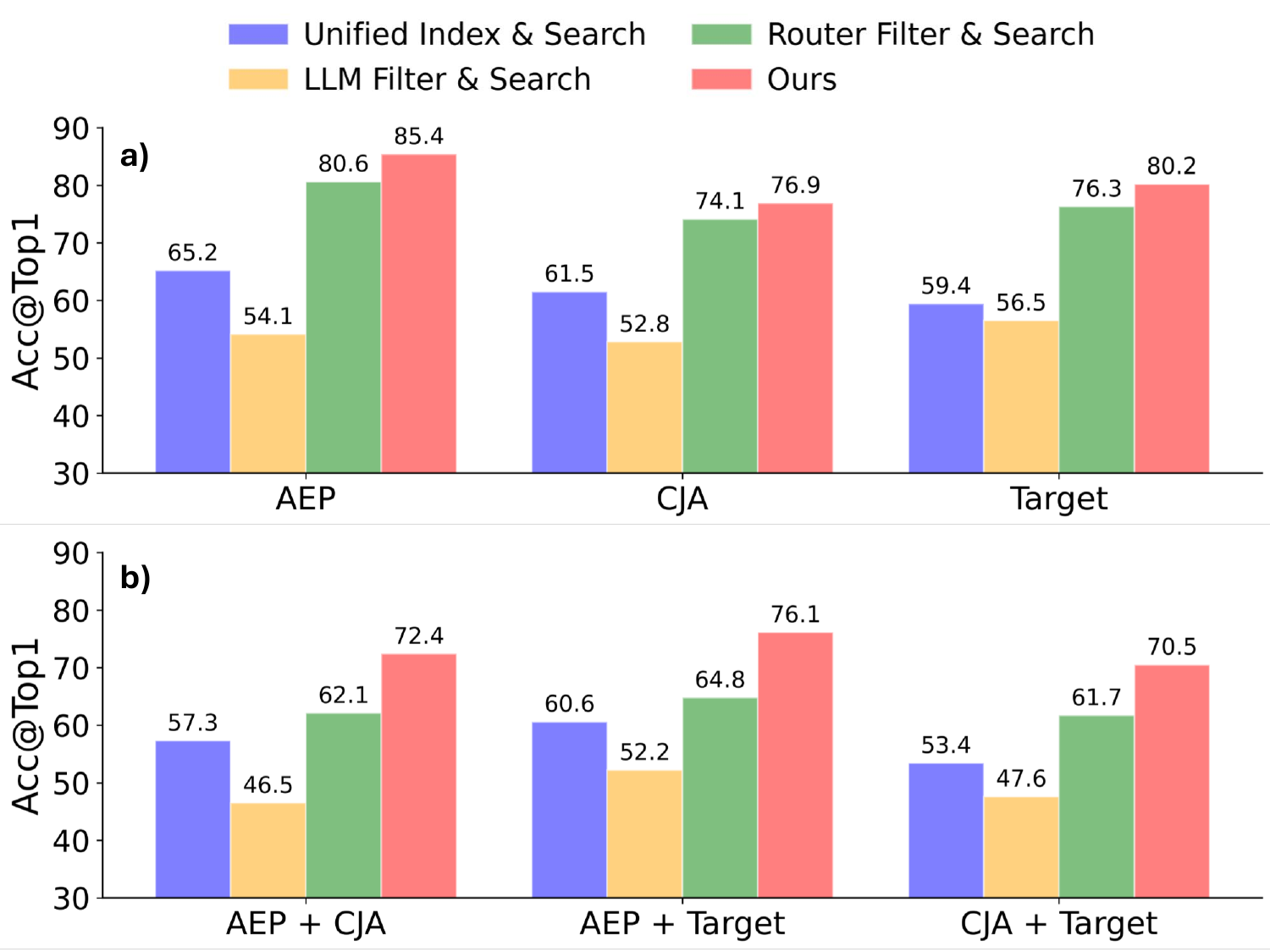}
\caption{
Performance comparison of retrieval accuracy (\text{Top-$1$}) across methods on \textbf{(a)} uni-domain, and \textbf{(b)} cross-domain datasets. 
}
\label{fig:acctop1}
\end{figure}

\begin{figure}[t]
\centering
\includegraphics[width=\columnwidth]{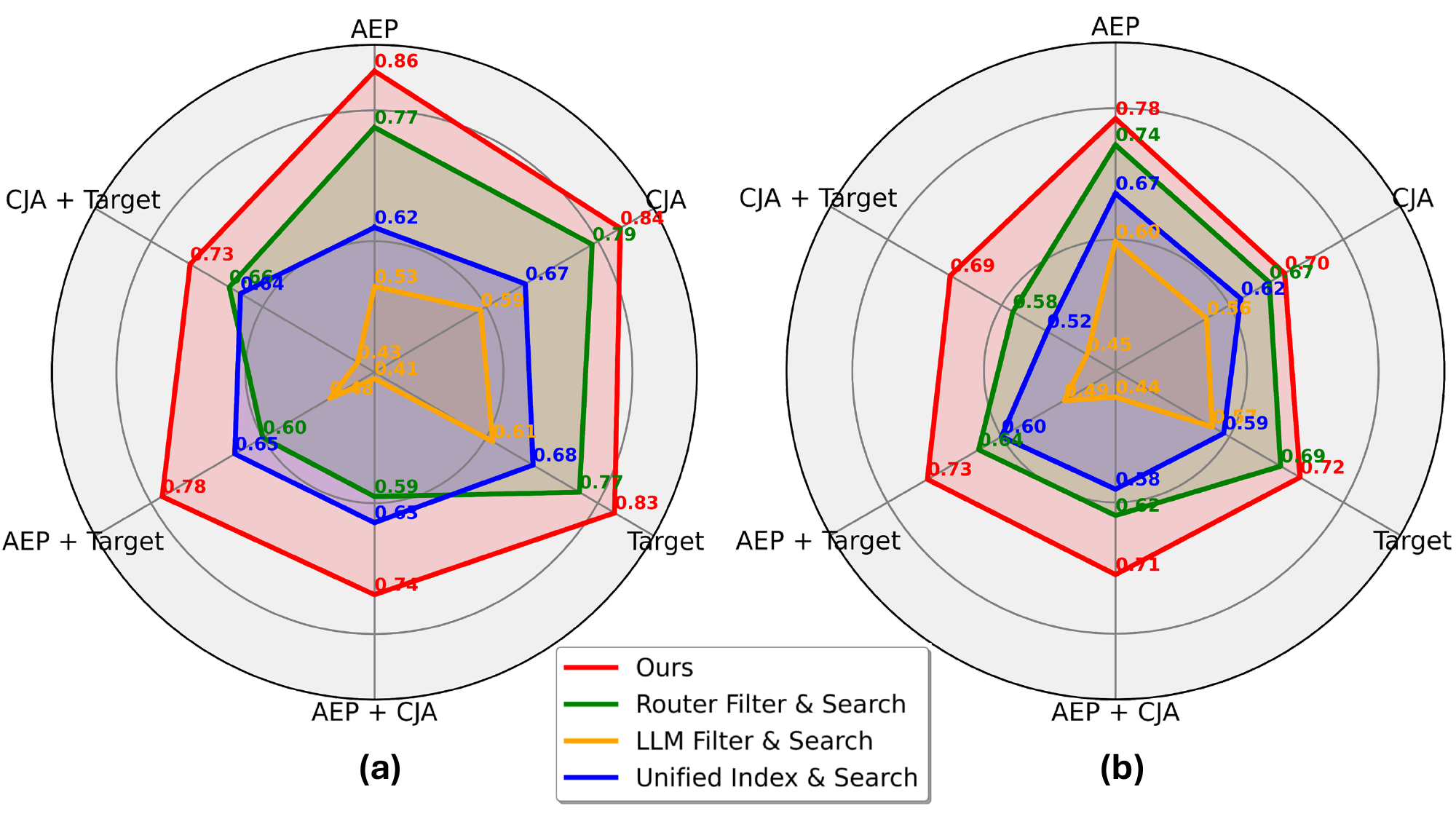}
\caption{
Performance comparison of response quality across methods on different datasets for LLM-based \textbf{(a)}~Relevancy, and \textbf{(b)}~ Faithfulness  metrics.
}
\label{fig:resp}
\end{figure}

\subsection{Results and Analysis}

\paragraph{Retrieval Performance}
Fig.~\ref{fig:acctop1} illustrates the retrieval accuracy (Acc@Top1) of our method and baselines across uni-domain and cross-domain datasets. Our approach consistently outperforms all baselines, with the performance gap widening in cross-domain settings. The RFS baseline shows the strongest performance among alternatives, particularly in uni-domain scenarios.  This can be attributed to the simpler query-domain relevance in uni-domain settings.
Conversely, the LFS baseline underperforms in retrieval accuracy, likely due to challenges general-purpose LLMs (e.g. GPT-4) face in domain selection and learning query-domain relevance for our specific product domains. 


\paragraph{Response Performance}
Figure~\ref{fig:resp} presents the assistant's response quality evaluations across all methods and datasets, focusing on Relevancy and Faithfulness metrics. 
Our method outperforms baselines on both metrics, with its advantage becoming more pronounced in cross-domain settings. 
The RFS baseline consistently ranks second in response quality, while the LFS baseline performs poorest.
Interestingly, the UIS baseline occasionally outperforms RFS on Relevancy in cross-domain datasets, but underperforms on Faithfulness. 
 This can be attributed to the fact that faithfulness correlates more strongly with retrieval accuracy, while relevancy assessment is usually influenced more by other factors like response length and context diversity which is prevalent in the UIS baseline setting.
\section{Path to Deployment}
Deploying \modelname into production requires careful planning, extensive testing, and significant efforts across multiple dimensions. We highlight the following key aspects that are essential for ensuring a successful and robust deployment and discuss our plans: 

\paragraph{Knowledge Precision}
 Accurate multi-domain federated search is crucial for retrieving relevant content across product domains, as inaccurate retrievals in RAG can lead to irrelevant or misleading AI assistant responses. Our goal is to achieve a retrieval accuracy (Acc@Top1) of 90\% or higher across both uni-domain and cross-domain queries. To improve this, we plan to implement regular retraining cycles for both the domain router and retriever models to adapt to evolving product documentation and user query patterns.

\paragraph{Latency}
Given the multi-step nature of our framework, managing response time is critical for user experience. Our target is to keep the end-to-end response time under 10 seconds for 95\% of queries. To do so, we plan to implement parallel processing for domain routing and document retrieval; utilize caching for frequently accessed documents; and explore quantization techniques for the retriever model to reduce inference time without significant accuracy loss.

\paragraph{User Study} A comprehensive user study is essential to evaluate performance in improving actual product QA experience. Once the system meets our accuracy and latency criteria, we plan to conduct A/B testing with a representative sample of users across different Adobe products, then, gather and analyze explicit and implicit user feedback through user surveys and interaction metrics (e.g., follow-up questions, task completion rates).

\paragraph{Continuous Monitoring and Iteration} 
Following the deployment of this work, continuous performance monitoring and iterative improvements are essential. We intend to implement monitoring dashboards tracking key performance metrics across different product domains; and establish a feedback loop where user interactions and support team insights are regularly incorporated into model retraining and system refinement.
\section{Conclusion}
In this paper, we introduced \modelname, a novel multi-domain knowledge-augmented question-answering framework for complex enterprise software ecosystems. Leveraging federated search with stochastic gating, \modelname outperforms baselines in retrieval accuracy and response quality across uni-domain and cross-domain settings for Adobe's Experience Platform (AEP), Target, and Customer Journey Analytics (CJA) applications.  
We also introduced new datasets for multi-product QA, addressing the lack of suitable benchmarks in this domain. Our findings highlight the importance of multi-domain knowledge integration and specialized approaches for domain-specific nuances in enterprise product QA, while also revealing limitations of LLM-based domain selection techniques.
Looking ahead, there are several avenues for future work and deployment optimization. These include implementing retraining cycles for router and retriever, exploring advanced caching and quantization techniques to reduce latency, and conducting comprehensive user studies to ensure alignment with real-world usage patterns.

\newpage

\bibliography{main_coling}

\appendix

\newpage
\clearpage

\section*{Appendix}
\label{sec:appendix}

\section{Query-Document Pair Examples}
The following AEP query-document examples highlight the necessity for product domain-specific knowledge in providing accurate and detailed responses to the user's questions:


\begin{figure}[!ht]
\centering
\includegraphics[width=0.9\columnwidth]{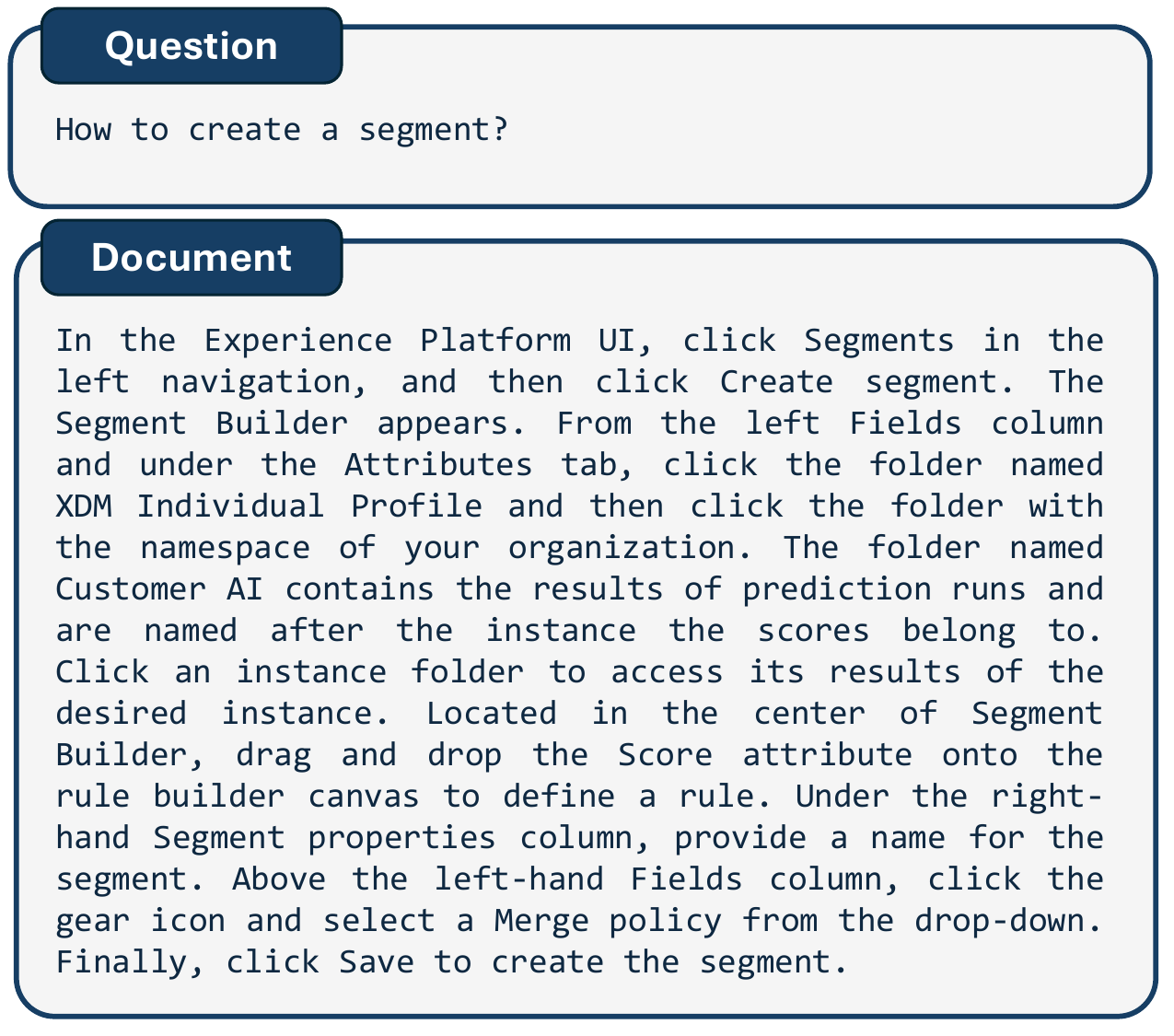}

\vspace{1.5em}
\includegraphics[width=0.9\columnwidth]{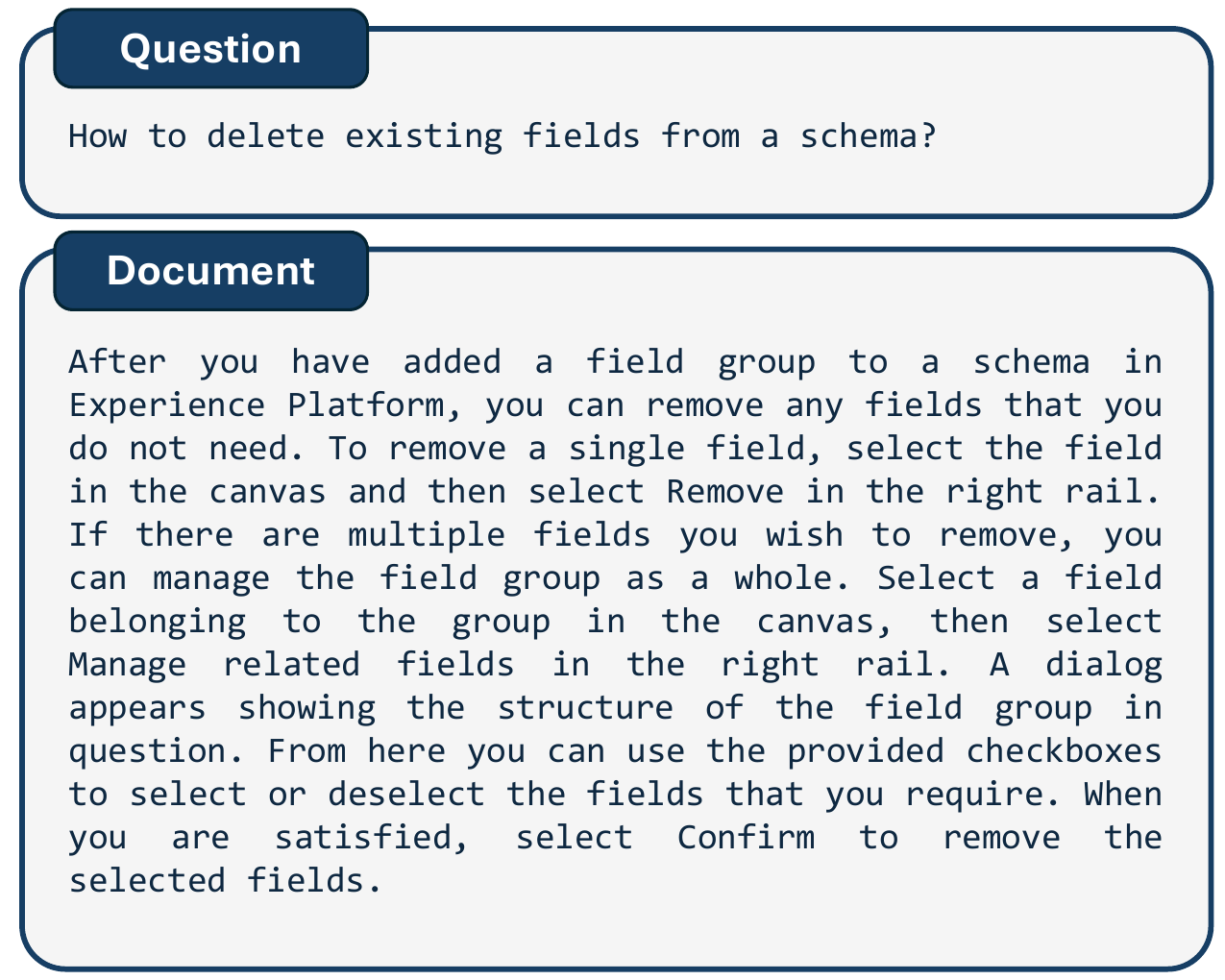}
\vspace{1.5em}

\includegraphics[width=0.9\columnwidth]{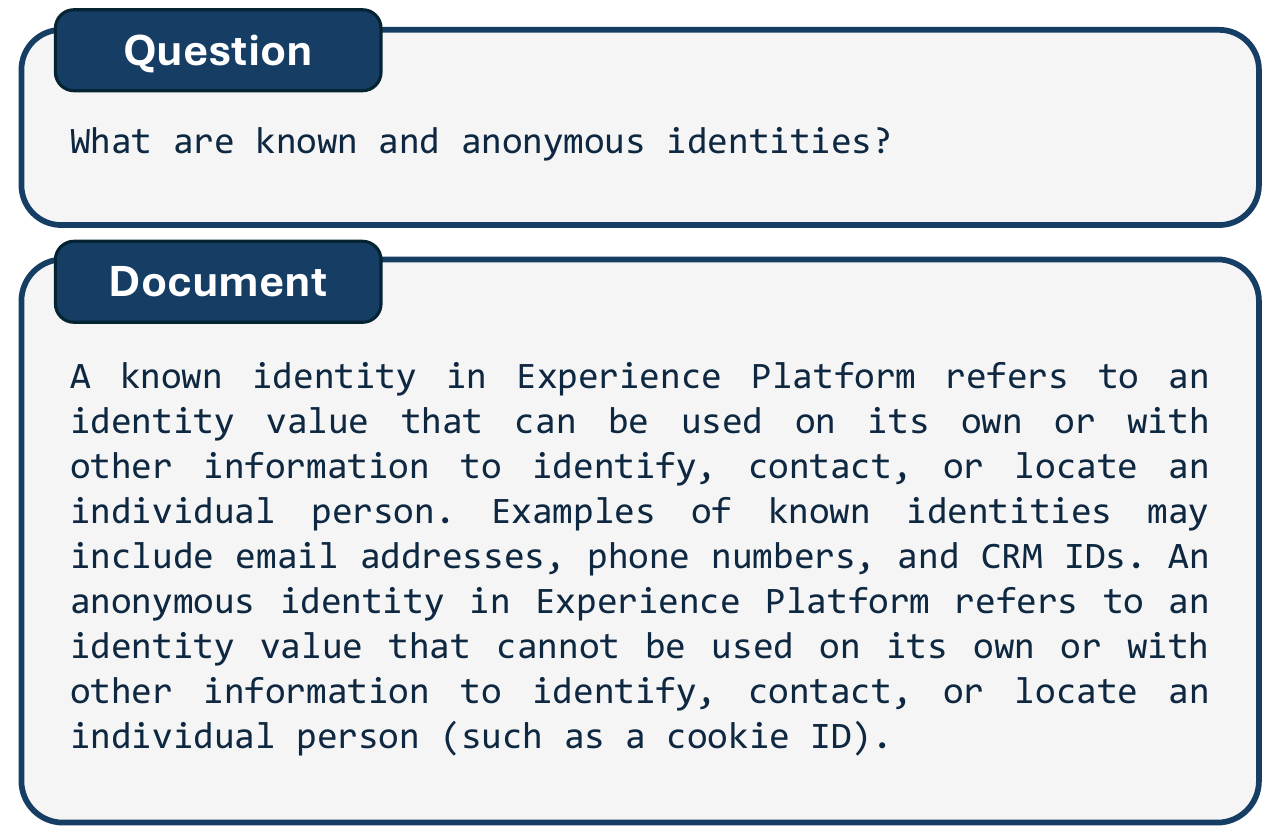}

\vspace{-0.5em}
\caption{
Examples of user query and relevant product documentation in the dataset.
}
\label{example-qd3}
\end{figure}

\section{LLM Prompts}
\label{sec:app-prompt}
This section provides an overview of the high-level structures for prompts utilized in our study



\begin{figure}[!ht]
\centering
\includegraphics[width=\columnwidth]{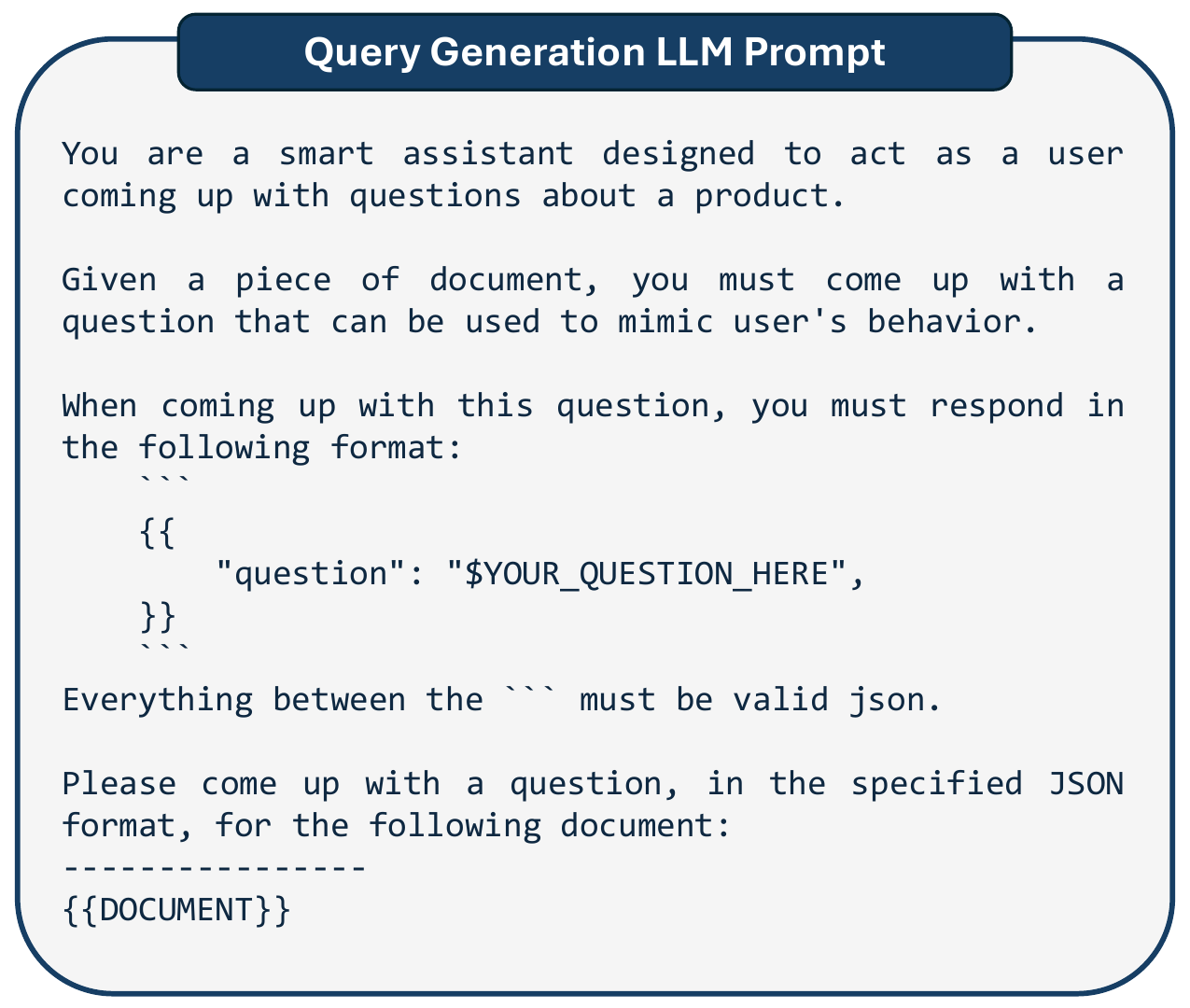}
\caption{
LLM Prompt for Query Generation: Simulating user behavior for document-based question synthesis 
}
\label{prompt-query}
\end{figure}

\begin{figure}[!ht]
\centering
\includegraphics[width=\columnwidth]{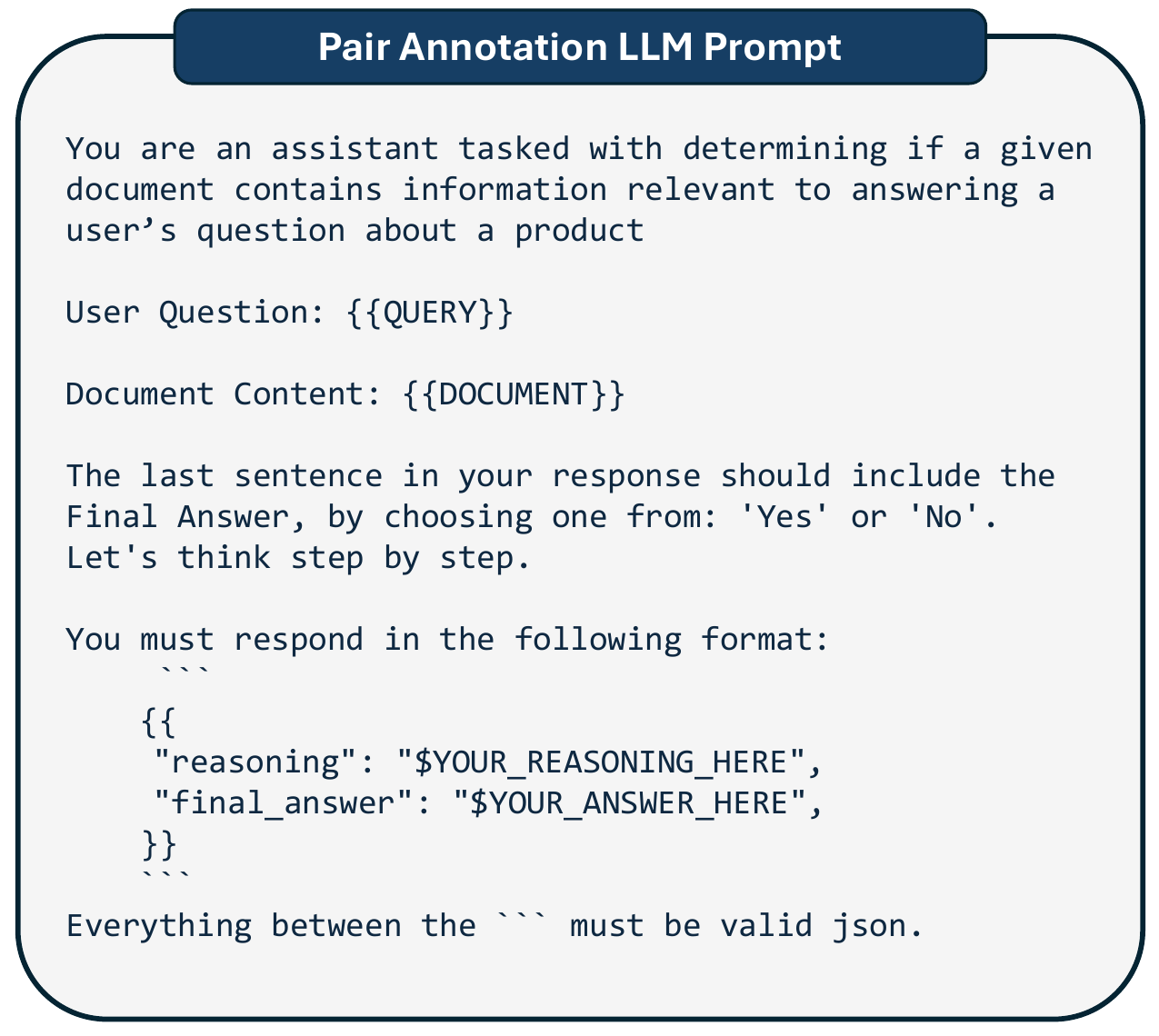}
\caption{
LLM Prompt for Query-Document Relevance Annotation: Binary labeling with explanatory reasoning for the relevance annotation.
}
\label{fig:prompt-annot}
\end{figure}

\begin{figure}[!ht]
\centering
\includegraphics[width=\columnwidth]{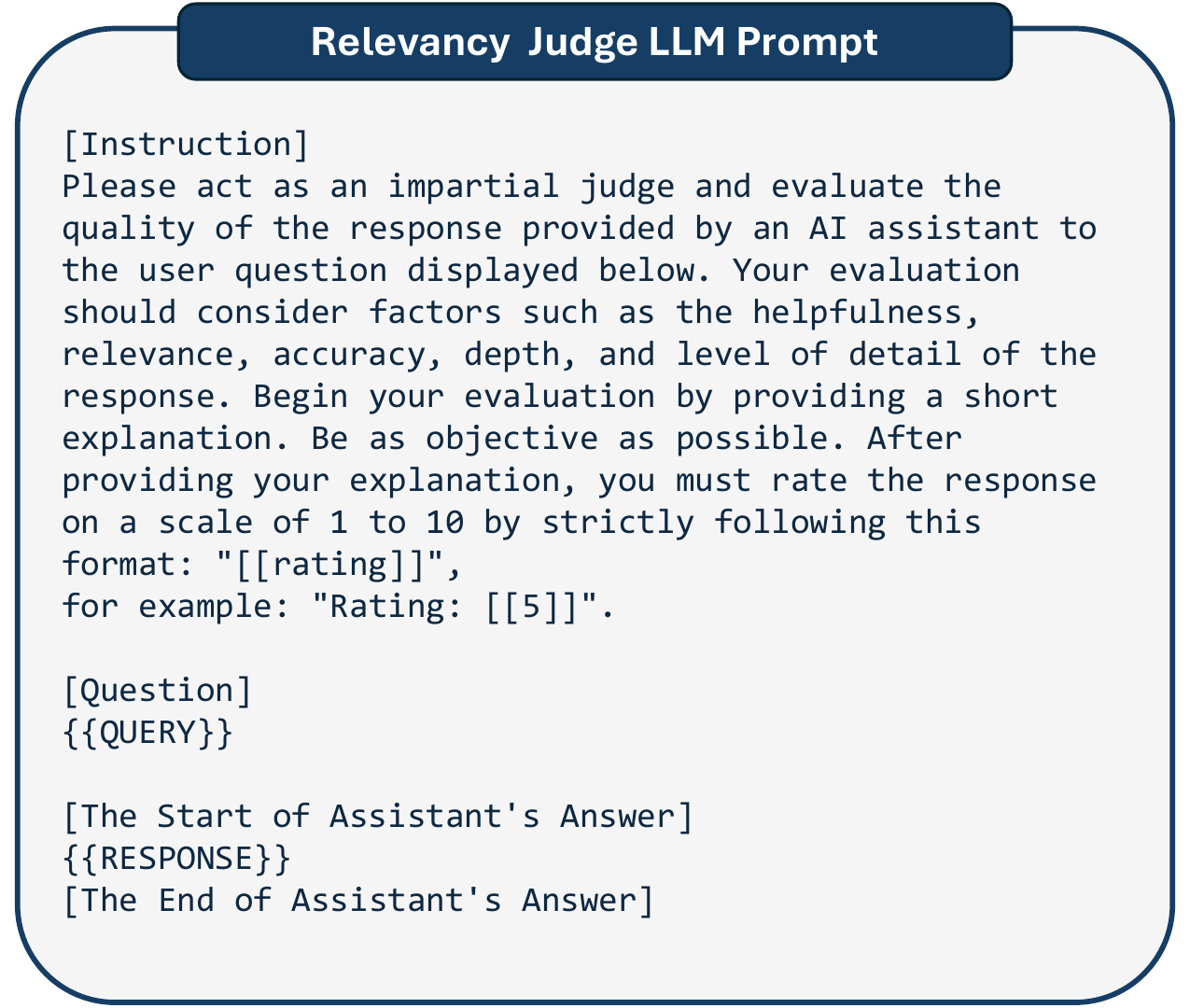}
\caption{
LLM Prompt for Query-Response Relevance Judge: Rating from 1 to 10 for the response to a given query,
following the prompt in \cite{zheng2024judging}.
}
\label{fig:prompt-relevancy}
\end{figure}

\begin{figure}[!ht]
\centering
\includegraphics[width=\columnwidth]{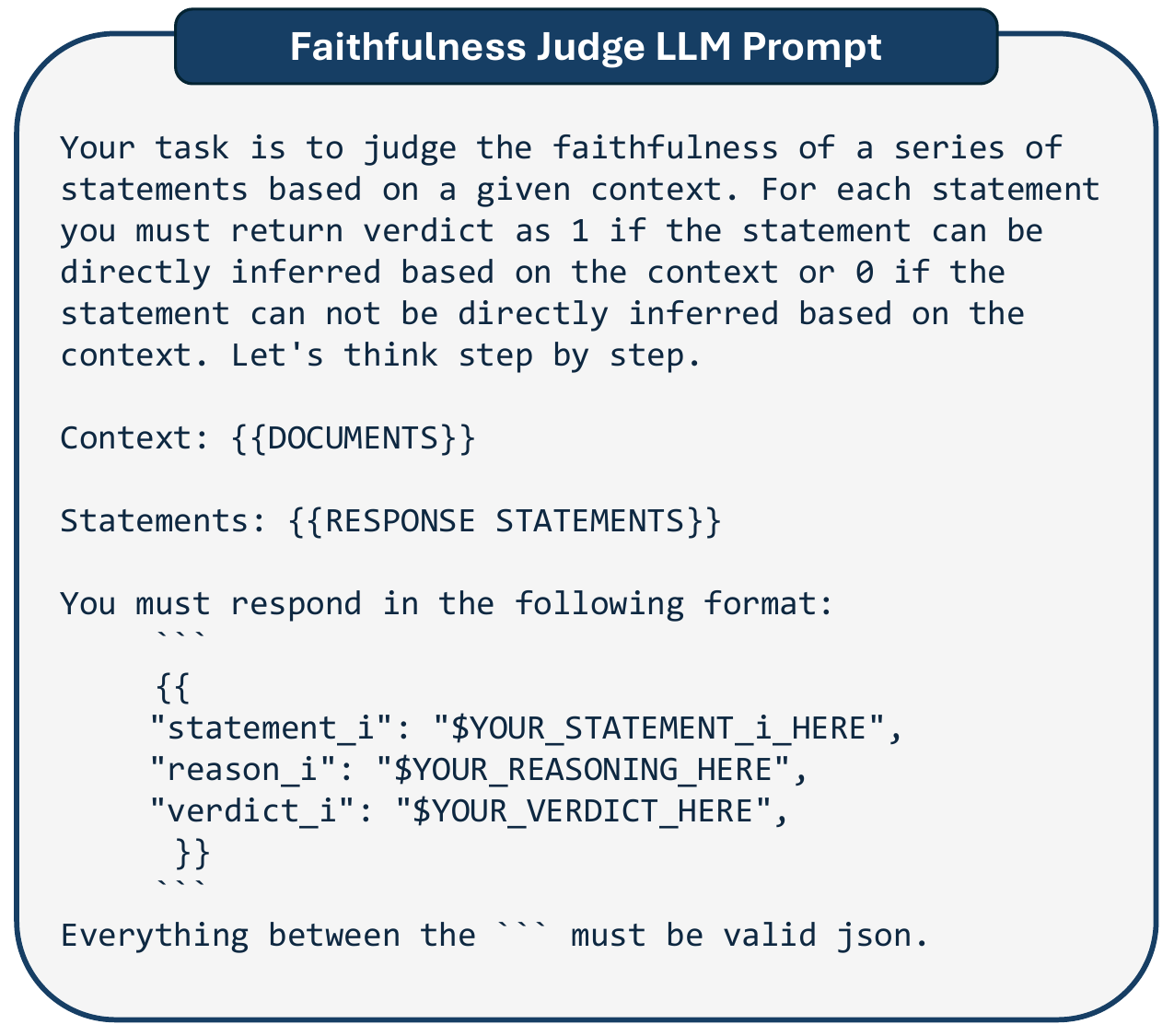}
\caption{
LLM Prompt for Response Faithfulness Judge: Binary rating for inferring each response statement from query's golden documents, 
following the prompt in \cite{es2023ragas}.
}
\label{fig:prompt-faithful}
\end{figure}

\begin{figure}[!ht]
\centering
\includegraphics[width=\columnwidth]{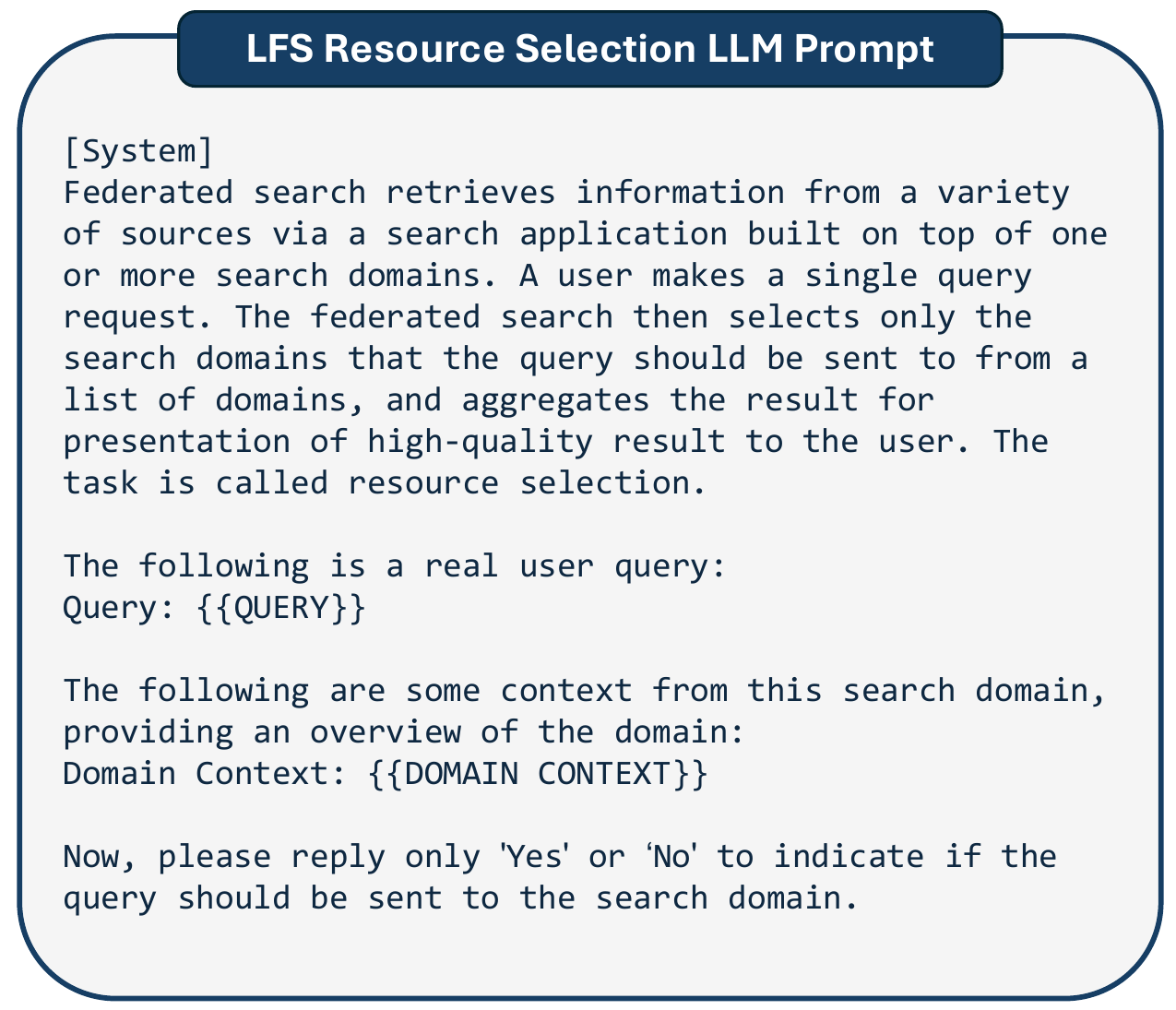}
\caption{
LLM Prompt for Resource Selection step in the LFS baseline,
following the prompt in \cite{wang2024resllm}.
}
\label{fig:prompt-resllm}
\end{figure}





\end{document}